\documentclass[letterpaper, 10 pt, conference]{class/ieeeconf}  % Comment this line out
\IEEEoverridecommandlockouts                              % This command is only
\usepackage{xcolor}
\usepackage{siunitx}  
\usepackage{graphicx}
 \graphicspath{./figures/robot_demonstration}
\usepackage{amsmath}
\usepackage{amssymb}
\usepackage{latexsym}
\usepackage{url}
\usepackage{cite}
\usepackage{relsize}
\usepackage{multirow}
\usepackage{afterpage}
\usepackage{ifthen}
\usepackage{graphicx}
\usepackage{algpseudocode,algorithm,algorithmicx}
\usepackage{subfigure}
\usepackage{flushend}
\usepackage{epstopdf}
\usepackage{stackengine}
\usepackage{textcomp} % new

\usepackage{gensymb}
\usepackage{booktabs}
\usepackage{makecell}
\usepackage{hhline}
\usepackage{courier}
\usepackage{lipsum}
\usepackage{xspace}
\usepackage{xcolor} % new
            
\makeatletter
\let\NAT@parse\undefined
\makeatother
\usepackage[bookmarks=false, linkcolor=blue, urlcolor=blue, citecolor=blue]{hyperref} 
\hypersetup{
    colorlinks=true,
    linkcolor=red,
    filecolor=magenta,      
    urlcolor=blue,
    pdfstartview={FitH},
    citecolor =blue
    }

\newcommand{\Real}{{\mathbb{R}}}

\newcommand{\yVIS}[1][]{y_{\text {vis}{#1}}}

\newcommand{\featVis}{\mathcal{F}^{\text vis}}
\newcommand{\yProb}{\rho}
\newcommand{\yPos}{r}
\newcommand{\yImgFeat}{y_{\text {img}}}

\usepackage{xspace}

\newcommand{\etal}{\textit{et al.}}

 \graphicspath{{./figures/}}

\title{\LARGE \bf Language-driven Grasp Detection with Mask-guided Attention}

\author{Tuan Van Vo$^{1}$, Minh Nhat Vu$^{2,3,*}$, Baoru Huang$^4$, An Vuong$^{1}$, Ngan Le$^5$, Thieu Vo$^6$, Anh Nguyen$^7$
\thanks{$^1$ FPT Software AI Center, Vietnam {\tt tuanvv7@fpt.com}}
\thanks{$^2$ Automation \& Control Institute, TU Wien, Austria %{\tt vu@acin.tuwien.ac.at}
}
\thanks{$^3$ Austrian Institute of Technology (AIT) GmbH, Austria  %{\tt vu@acin.tuwien.ac.at}
}
\thanks{$^4$ Imperial College London, UK
}
\thanks{$^5$ Department of Computer Science \& Computer Engineering, University of Arkansas, USA 
}
\thanks{$^6$ National University of Singapore, Singapore} %{\tt vongocthieu@tdtu.edu.vn}}
\thanks{$^7$ Department of Computer Science, University of Liverpool, UK %{\tt anh.nguyen@liverpool.ac.uk}
}
\thanks{$^*$ Corresponding author {\tt minh.vu@ait.ac.at}
}
}

\begin{document}
% Macros

\newtheorem{problem}{Problem}
\newtheorem{lemma}{Lemma}
\newtheorem{theorem}[lemma]{Theorem}
\newtheorem{claim}{Claim}
\newtheorem{corollary}[lemma]{Corollary}
\newtheorem{definition}[lemma]{Definition}
\newtheorem{proposition}[lemma]{Proposition}
\newtheorem{remark}[lemma]{Remark}
\newenvironment{LabeledProof}[1]{\noindent{\it Proof of #1: }}{\qed}

\def\beq#1\eeq{\begin{equation}#1\end{equation}}
\def\bea#1\eea{\begin{align}#1\end{align}}
\def\beg#1\eeg{\begin{gather}#1\end{gather}}
\def\beqs#1\eeqs{\begin{equation*}#1\end{equation*}}
\def\beas#1\eeas{\begin{align*}#1\end{align*}}
\def\begs#1\eegs{\begin{gather*}#1\end{gather*}}

\newcommand{\poly}{\mathrm{poly}}
\newcommand{\eps}{\epsilon}
\newcommand{\e}{\epsilon}
\newcommand{\polylog}{\mathrm{polylog}}
\newcommand{\rob}[1]{\left( #1 \right)} %Round Brackets
\newcommand{\sqb}[1]{\left[ #1 \right]} %square Brackets
\newcommand{\cub}[1]{\left\{ #1 \right\} } %curly brackets
\newcommand{\rb}[1]{\left( #1 \right)} %Round
\newcommand{\abs}[1]{\left| #1 \right|} %| |
\newcommand{\zo}{\{0, 1\}}
\newcommand{\zonzo}{\zo^n \to \zo}
\newcommand{\zokzo}{\zo^k \to \zo}
\newcommand{\zot}{\{0,1,2\}}
\newcommand{\en}[1]{\marginpar{\textbf{#1}}}
\newcommand{\efn}[1]{\footnote{\textbf{#1}}}
\newcommand{\vecbm}[1]{\boldmath{#1}} %more general (handles greek letters)
\newcommand{\uvec}[1]{\hat{\vec{#1}}}
\newcommand{\thv}{\vecbm{\theta}}
\newcommand{\junk}[1]{}
\newcommand{\var}{\mathop{\mathrm{var}}}
\newcommand{\rank}{\mathop{\mathrm{rank}}}
\newcommand{\diag}{\mathop{\mathrm{diag}}}
\newcommand{\tr}{\mathop{\mathrm{tr}}}
\newcommand{\acos}{\mathop{\mathrm{acos}}}
\newcommand{\atantwo}{\mathop{\mathrm{atan2}}}
\newcommand{\SVD}{\mathop{\mathrm{SVD}}}
\newcommand{\quadf}{\mathop{\mathrm{q}}}
\newcommand{\linterp}{\mathop{\mathrm{l}}}
\newcommand{\sgn}{\mathop{\mathrm{sign}}}
\newcommand{\sym}{\mathop{\mathrm{sym}}}
\newcommand{\avg}{\mathop{\mathrm{avg}}}
\newcommand{\mean}{\mathop{\mathrm{mean}}}
\newcommand{\erf}{\mathop{\mathrm{erf}}}
\newcommand{\grad}{\nabla}
\newcommand{\R}{\mathbb{R}}
\newcommand{\defeq}{\triangleq}
\newcommand{\dims}[2]{[#1\!\times\!#2]}
\newcommand{\sdims}[2]{\mathsmaller{#1\!\times\!#2}}
\newcommand{\udims}[3]{#1}
\newcommand{\udimst}[4]{#1}
\newcommand{\com}[1]{\rhd\text{\emph{#1}}}
\newcommand{\ind}{\hspace{1em}}
\newcommand{\argmin}[1]{\underset{#1}{\operatorname{argmin}}}
\newcommand{\floor}[1]{\left\lfloor{#1}\right\rfloor}
\newcommand{\step}[1]{\vspace{0.5em}\noindent{#1}}
\newcommand{\quat}[1]{\ensuremath{\mathring{\mathbf{#1}}}}
\newcommand{\norm}[1]{\left\lVert#1\right\rVert}
\newcommand{\ignore}[1]{}
\newcommand{\specialcell}[2][c]{\begin{tabular}[#1]{@{}c@{}}#2\end{tabular}}
\newcommand*\Let[2]{\State #1 $\gets$ #2}
\newcommand{\algorithmicbreak}{\textbf{break}}
\newcommand{\Break}{\State \algorithmicbreak}
\newcommand{\ra}[1]{\renewcommand{\arraystretch}{#1}}

\renewcommand{\vec}[1]{\mathbf{#1}} %looks better

\algdef{S}[FOR]{ForEach}[1]{\algorithmicforeach\ #1\ \algorithmicdo}
\algnewcommand\algorithmicforeach{\textbf{for each}}
\algrenewcommand\algorithmicrequire{\textbf{Require:}}
\algrenewcommand\algorithmicensure{\textbf{Ensure:}}
\algnewcommand\algorithmicinput{\textbf{Input:}}
\algnewcommand\INPUT{\item[\algorithmicinput]}
\algnewcommand\algorithmicoutput{\textbf{Output:}}
\algnewcommand\OUTPUT{\item[\algorithmicoutput]}

\maketitle
\thispagestyle{empty}
\pagestyle{empty}

%%%%%%%%%%%%%%%%%%%%%%%%%%%%%%%%%%%%%%%%%%%%%%%%%%%%%%%%%%%%%%%%%%%%%%%%%%%%%%%%
\begin{abstract}
Grasp detection is an essential task in robotics with various industrial applications. However, traditional methods often struggle with occlusions and do not utilize language for grasping. Incorporating natural language into grasp detection remains a challenging task and largely unexplored. To address this gap, we propose a new method for language-driven grasp detection with mask-guided attention by utilizing the transformer attention mechanism with semantic segmentation features. Our approach integrates visual data, segmentation mask features, and natural language instructions, significantly improving grasp detection accuracy. Our work introduces a new framework for language-driven grasp detection, paving the way for language-driven robotic applications. Intensive experiments show that our method outperforms other recent baselines by a clear margin, with a $10.0\%$ success score improvement. We further validate our method in real-world robotic experiments, confirming the effectiveness of our approach. 
\end{abstract}

%%%%%%%%%%%%%%%%%%%%%%%%%%%%%%%%%%%%%%%%%%%%%%%%%%%%%%%%%%%%%%%%%%%%%%%%%%%%%%%%

\section{INTRODUCTION} \label{Sec:Intro}
%DO NOT DELETE \note{figure link: shorturl.at/glALZ} 
Grasp detection is the fundamental task in robotics, with widespread applications in manufacturing, logistics, and service robots~\cite{sun2021research}. Traditional grasping detection methods often struggle with object complexities and occlusions~\cite{gilles2023metagraspnetv2}. However, recent advances in computer vision, machine learning, and natural language processing have opened up new possibilities for addressing the challenge using deep networks~\cite{shridhar2022cliport}. However, most existing works focus on detecting grasp poses without language instruction~\cite{maitin2010cloth, domae2014fast,nguyen2016preparatory,jiang2011efficient, depierre2018jacquard,fang2020graspnet,wang2022transformer}. In practice, language-driven grasping presents an intriguing and demanding task in robot manipulation~\cite{shridhar2022cliport, vuong2023grasp}, where natural language can guide the robot to grasp on-demand objects. Developing language-driven grasping systems is not a rival task and requires the understanding of language instructions and visual information of scene~\cite{platt2023grasp, song2023llm}.

In recent years, there has been a surge in interest in language-driven robotic manipulation, enabling robots to comprehend natural language commands for executing manipulation tasks~\cite{nguyen2023open,van2023open,shridhar2022cliport,driess2023palm,platt2023grasp}. This paradigm shift brings numerous advantages, including enhanced human-robot interaction, adaptability to various environments, and improved task efficiency~\cite{mu2024embodiedgpt}. Language-driven robotic frameworks are gaining momentum, empowering robots to process natural language and bridging the gap between robotic manipulations and real-world human-robot interaction~\cite{mu2024embodiedgpt}. Embodied robots such as PaLM-E~\cite{driess2023palm}, EgoCOT~\cite{mu2024embodiedgpt}, and ConceptFusion~\cite{jatavallabhula2023conceptfusion} have emerged with the capability to comprehend natural language by leveraging large foundation models like ChatGPT~\cite{chatgpt}. However, many existing works primarily focus on high-level robot actions, overlooking fundamental grasping actions, thus limiting generalization across robotic domains, tasks, and skills~\cite{mu2023ec2}. Despite the rising interest in language-driven grasp detection, current approaches struggle to handle object complexities effectively~\cite{mu2023ec2}. Challenges such as ambiguities in natural language instructions, limited vocabulary, and difficulties in contextual understanding impede the accurate interpretation of user commands~\cite{mu2024embodiedgpt}. Moreover, dependencies on precise language understanding and inefficiencies in noisy environments pose additional obstacles, potentially leading to difficulties in object comprehension~\cite{platt2023grasp}.

\begin{figure}
\centering
\includegraphics[width=0.99\linewidth]{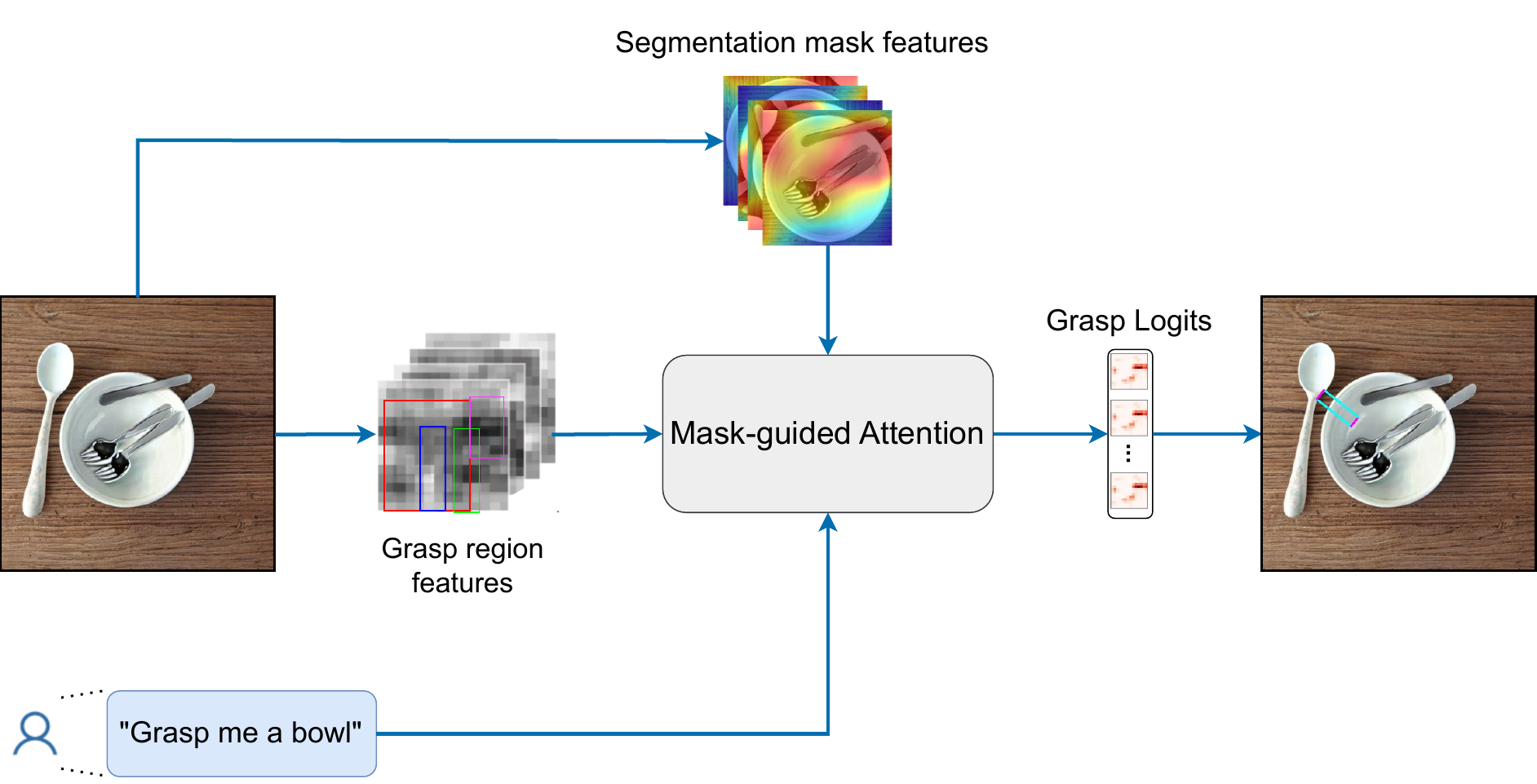}
\vspace{0.5ex}
\caption{We propose a mask-guided attention mechanism that learns the mask and language features to tackle the language-driven grasping task.}
\label{fig:intro}
\end{figure}

In this paper, we present a new approach to tackle the language-driven grasp detection task. Inspired by the Transformer network's powerful attention mechanism~\cite{devlin2018bert}, our method capitalizes on recent advancements in multimodal learning to integrate visual information, segmentation mask features, and natural language instructions for robust grasp detection. Specifically, we propose a mask-guided attention mechanism for the language-driven grasp detection task to concurrently model grasp region features, segmentation mask features, and language embeddings. Our approach aims to enhance object understanding through attention to segmentation mask features, facilitating a better connection between attended language embeddings and the correct grasp region features, thereby improving the accuracy of the language-driven grasp detection task. Extensive experiments demonstrate that our method achieves an approximately $10\%$ success score improvement over the baselines. Ablation studies and qualitative results on real-world robotic grasping applications further validate the effectiveness of our approach and provide insights for future research directions.

Our main contributions are summarized as follows:
\begin{itemize}
    \item We propose a mask-guided attention mechanism to enhance multimodal integration for the language-driven grasp detection task. %It improves object understanding by emphasizing semantic segmentation features and enhancing the connection between language embeddings and objects.
    \item We provide a comprehensive analysis of our proposed method, including experimental results on benchmark datasets and ablation studies to evaluate the effectiveness of different components. % Our code will be made available.
    %\item We also compare our approach with existing state-of-the-art methods in language-driven grasp detection, demonstrating superior performance in terms of grasp accuracy, adaptability to clutter, and robustness to occlusions.    
\end{itemize}

\section{Related Work} \label{Sec:rw}

% As the field continues to evolve, alternative approaches are being explored, ranging from lightweight language representations to domain-specific language understanding, aiming to address the challenges and limitations associated with current language-driven grasp detection methodologies.
\textbf{Grasp Detection.} Traditional approaches to robotic grasp detection have included analytical methods~\cite{maitin2010cloth, domae2014fast, roa2015grasp}, which focus on object geometry and contact forces, and convolutional neural networks (CNNs)~\cite{morrison2020learning, kumra2020antipodal, jiang2011efficient,lenz2015deep, pinto2016supersizing, mahler2017dex}, trained on labeled datasets of grasping examples~\cite{jiang2011efficient, depierre2018jacquard, fang2020graspnet}. While attention mechanisms in Transformers have shown promise in sequence modeling for information fusion across global sequences~\cite{devlin2018bert}, Wang~\etal~\cite{wang2022transformer} introduced a Transformer-based visual grasp detection framework, leveraging attention's ability to aggregate information across input sequences for improved global representation. The design of this framework incorporates local window attention to capture local contextual information and detailed features of graspable objects. However, a significant drawback of both analytical, CNN-based, and Transformer-based methods is their limited scene understanding and inability to process language instructions, which hampers their effectiveness in dynamic, human-centric environments.

\textbf{Language-driven Grasp Detection}. Recent advances in large language models have facilitated the integration of language understanding to robotic tasks, enabling robots to execute manipulation tasks based on natural language instructions ~\cite{lu2023vl, sun2023language, nguyen2024LGrasp6D, cheang2022learning, vuong2024language,nghia2024fastgrasp, xu2023joint, yang2022interactive}. This transition towards language-guided grasp referral empowers robots to identify and manipulate objects according to user specifications, thereby enhancing their utility in complex scenarios~\cite{mu2024embodiedgpt,platt2023grasp}. For instance, Tziafas~\etal~\cite{tziafas2023language} concentrate on grasp synthesis based on linguistic references, predicting grasp poses for referenced objects using natural language in cluttered scenes, while Chen~\etal~\cite{chen2021joint} propose a method to jointly learn from visual and language features and predict 2D grasp boxes from RGB images.

\textbf{Transformer Attention Mechanism.} Initially for NLP tasks~\cite{vaswani2017attention}, the Transformer's multi-head attention mechanism excels in capturing long-term word correlations. While primarily used in NLP, attempts have been made to apply Transformers to vision tasks such as image super-resolution~\cite{yang2020learning}, object detection~\cite{carion2020end}, and multimodal video understanding~\cite{sun2019learning, chen2020uniter, luo2020univl}. However, these methods still rely on CNN-extracted features. Recent advancements include convolution-free vision Transformers~\cite{dosovitskiy2021an}, which operate directly on raw images, achieving competitive performance. Further improvements in training data efficiency have been made by \cite{touvron2020training} through stronger data augmentations and knowledge distillation. The pure Transformer design has since been applied to various vision tasks, including semantic segmentation~\cite{zheng2020rethinking}, point cloud classification~\cite{zhao2020point}, and action recognition~\cite{bertasius2021space, sharir2021image, vivit}. In our work, we propose Mask-guided Attention as the Transformer attention model to learn visual inputs, object segmentation features, and text features for the language-driven grasp detection task.

Despite the burgeoning interest in language-driven grasp detection, extant methods grapple with the intricacies of object geometries, linguistic ambiguities, and contextual understanding challenges, impeding precise interpretation of user commands~\cite{mu2024embodiedgpt, driess2023palm}. This restricts robots' ability to understand nuanced, implicit instructions crucial for real-world interactions~\cite{mu2023ec2, mu2024embodiedgpt, platt2023grasp}. To this end, we introduce a new framework for language-driven grasp detection, harnessing recent strides in transformer multimodal learning and attention mechanisms~\cite{devlin2018bert}. Specifically, we advocate for a mask-guided attention mechanism, tailored to bolster grasp detection reliability through comprehensive multimodal integration. Our intuition is to utilize segmentation mask features to enhance the alignment between language embeddings and grasp region features for the language-driven grasping task.

% Language-driven grasp detection has gained substantial traction in the field, representing a significant intersection of computer vision and robotics~\cite{shridhar2022cliport, fang1billion}. While conventional grasp detection methods have leaned on geometric and visual features~\cite{ainetter2021end, depierre2018jacquard, fang1billion, shridhar2022cliport}, recent progress highlights the integration of natural language to enrich human-robot interaction~\cite{shridhar2022cliport, xu2023joint, yang2023pave}. However, prevalent approaches tend to concentrate on higher-level robot actions, sidelining essential grasping actions, and consequently constraining generalization across robotic domains, tasks, and skills~\cite{mu2023ec2}.  
\section{Language-driven Grasping with Mask-guided Attention} \label{Sec:method}

\begin{figure*}[t]
	\centering	\includegraphics[width=1.0\linewidth]{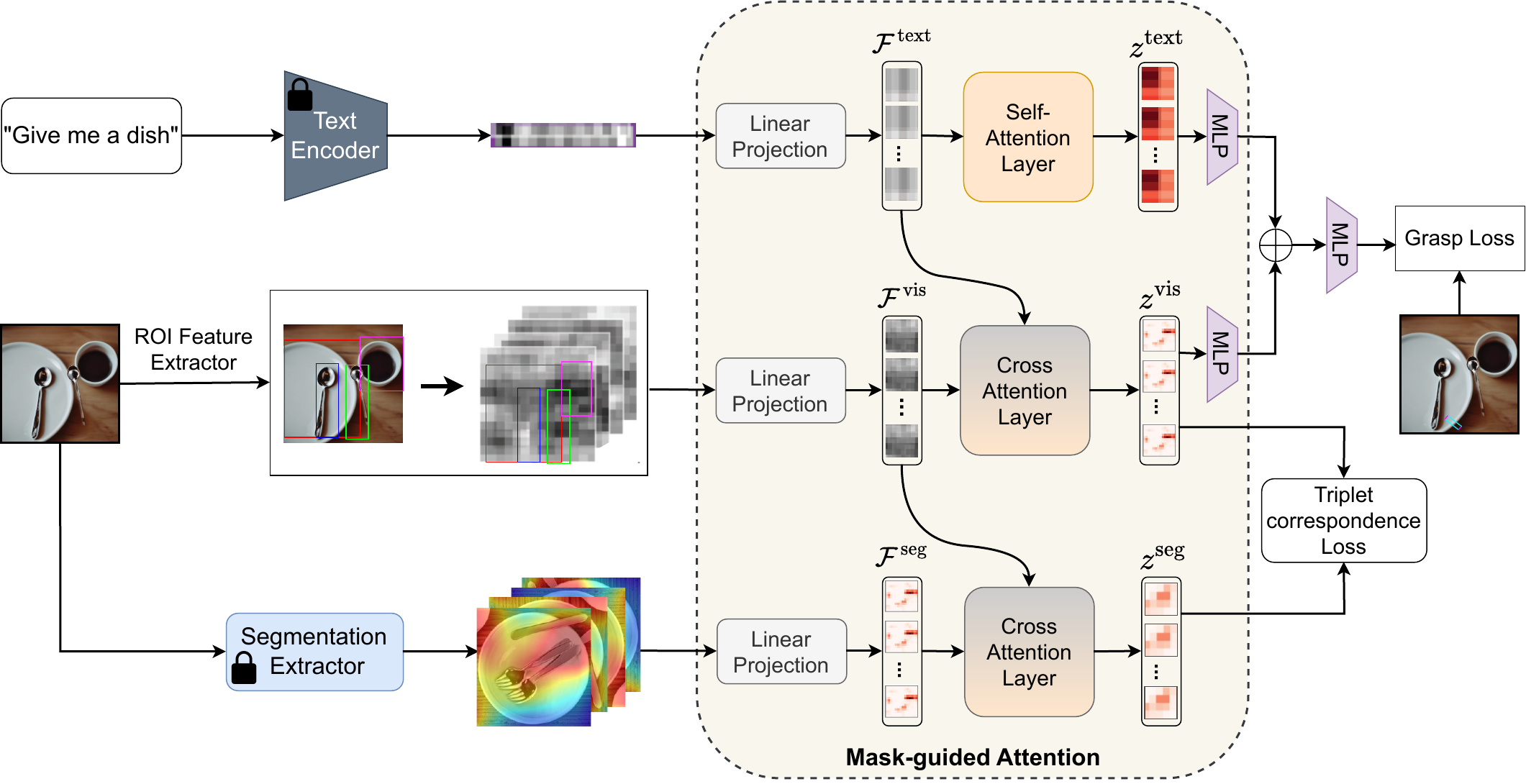}
 \vspace{1pt}
	\caption{The overview of our mask-guided attention framework for the language-driven grasp detection task.}
	\label{fig:architecture}
\end{figure*}

\subsection{Overview}
Given an input RGB image $I$ and a text prompt describing the object of interest, our goal is to detect the grasping pose on the image that best matches the text prompt input. We follow the popular \textit{rectangle grasp} convention that is widely used in previous work to define the grasp pose~\cite{depierre2018jacquard}. In particular, each grasp pose is defined with five parameters: the $(x,y)$ center coordinate, the width, height $(w,h)$ of the rectangle, and the rotational angle identifies the orientation of the rectangle relative to the horizontal axis of the image. Fig.~\ref{fig:architecture} shows an overview of our framework. We leverage a pretrained text encoder and segmentation mask features from the segmentation head, along with a grasp region proposal head for spatial feature extraction. Our method, mask-guided attention, improves grasp detection task by incorporating cross-attention from segmentation mask features and language embeddings, enhancing the connection between language embeddings and grasp region features through attention to segmentation mask features.

\subsection{Visual and Language Feature Extraction}

\textbf{Grasp Region Feature Extraction.} The first stage of our visual processing pipeline is to extract a set of grasp
regions of interest (ROIs) and their feature representations as in~\cite{chu2018real,roiGrasp2019}. The visual grasp region features contain geometric information for determining grasp configuration and semantic information for reasoning with natural languages. At the end of the grasp region feature extraction pipeline, a fixed-size feature map is passed to a convolution neural network
to produce a set of vectors% in $\Real^\dimImg$ \textcolor{blue}{updated} \textbf{???}
, which are interpreted as the embedding for each candidate grasp region. Specially, the grasp region feature representation vector is defined as: 
%The output is set of such vectors, \textbf{CHECK THE EQUATION, SEEMS NOT OK}
\begin{equation} 
\begin{split} 
  \featVis & = \{ \yVIS[,i] \}_{1}^{m} 
           = \{(\yProb^i, \yPos^i, \yImgFeat^i)\}_{1}^{m},
%  \\
%  \fVis & = \fVEC \circ \fROIPOOL \circ \fRPN \circ \fENC,
\end{split}
\end{equation}
where $\featVis \in \Real^{d\times h\times w}$, the coordinates of $(\yProb, \yPos, \yImgFeat) $
consist of the proposal probability, the predicted position, and the
image feature representation.

%\textcolor{blue}{Answer: Let me take a look. But we train this by the grasp grouth by L1 Loss (like regression), and Maybe better if we have explaiend in the Training Loss} \textbf{ADD 1 OR 2 SENTENCE HOW TO TRAIN, AND Groundtruth need for this step? or unsupervised?}

\textbf{Segmentation Feature Extraction.} We leverage an object instances segmentation network to acquire features that represent the ``meaning" of objects within an image. By building upon the proven object instance segmentation architecture~\cite{xiang2021learning,back2022unseen}, our network progressively analyzes the image, culminating in rich, high-dimensional segmentation mask features $\mathcal{F}^{\text{seg}} \in \Real^{d\times h\times w}$. These features provide a granular understanding of the scene, from large shapes to fine details, leading to improved performance.

\textbf{Text embedding.}
Given an input text query with $K$ words (e.g., ``grasp the blue bottle"), we embed the text input with a pre-trained BERT~\cite{devlin2018bert} or CLIP~\cite{radford2021learning} into text embedding feature vectors $\mathcal{F}^\text{text}\in 
\Real^d$. We note that the text encoder is frozen during the training.

% %%% WE CAN SKIP THIS PARAGRAPH. THE THEORY IN HERE IS EXPLAINED LATER.
% \textbf{Grasp Conditioned-Mask Multimodal Cross Attention.} \textcolor{red}{todo}\textbf{DON'T KNOW WHERE IS IT IN THE FIGURE}
% After respectively embedding each modality into multiple $d$-dimension feature vectors, we apply a stack of transformer layers~\cite{devlin2018bert} to fuse the input modalities (query words, 2D grasp objects proposals, and  training object mask semantics). We denote the transformer's output features at the language, grasp proposal, and  segmentation object semantic features positions as $z^{t}$, $z^{vis}$, and $z^{seg}$. An output grounding module that consists of two fully connected layers projects fused features $\{z^{vis}_1,\cdots,z_M^{vis}\}$ into a set of $M$ grasp scores $\{S^{vis}_1,\cdots,S_M^{vis}\}$, respectively. The grasp object proposal $G_i, i \in M$ with the highest grasping score is selected as the final grasping prediction.

\subsection{Mask-guided Attention}
Inspired by the Transformer network's powerful attention mechanism~\cite{devlin2018bert}, we introduce a new architecture called mask-guided attention. Our approach merges information from various sources (grasp region features, text features, and segmentation mask features) to achieve a deeper understanding of grasping tasks. By employing cross-modal attention, our method focuses on critical features within each modality, ultimately fusing them into a unified representation that guides toward robust grasping. Our proposed cross attention mechanism jointly learns the $W^\text{text}_Q,W^\text{text}_K,W^\text{text}_V$ and $W^\text{vis}_Q,W^\text{vis}_K,W^\text{vis}_V$ and $W^\text{seg}_Q,W^\text{seg}_K,W^\text{seg}_V$, i.e., the query, key and value weight matrices for grasp region features, text features, and segmentation mask features, respectively. Here, all weight matrices have dimensions $d \times d$. 

We first use a self-attention layer to compute the output $S^\text{text}$ from the input features $\mathcal{F}^\text{text}$ by first transforming them into query, key, and value matrices using learned linear transformations. Specifically, we calculate $Q^\text{text}= \mathcal{F}^\text{text}\times W^\text{text}_Q$; $K^\text{text}= \mathcal{F}^\text{text}\times W^\text{text}_K$; $V^\text{text}= \mathcal{F}^\text{text}\times W^\text{text}_V$. Subsequently, we determine the value $S^\text{text}$ as follows:

\begin{equation}
\label{eq:text_att}
    S^\text{text}=\text{softmax}(\dfrac{Q^\text{text}\cdot (K^\text{text})^\mathrm{\top}}{\sqrt{d}})
\end{equation}

To understand the relationship between the text and grasp region features, we first calculate $Q^\text{vis}= \mathcal{F}^\text{text}\times W^\text{vis}_Q$; $K^\text{vis}= \mathcal{F}^\text{vis}\times W^\text{vis}_K$; $V^\text{vis}= \mathcal{F}^\text{vis}\times W^\text{vis}_V$, then calculate $S^\text{vis}$:
\begin{equation}
\label{eq:image_att}
    S^\text{vis}=\text{softmax}(\dfrac{Q^\text{vis}\cdot (K^\text{vis})^\mathrm{\top}}{\sqrt{d}})
\end{equation}

% For the grasp region features, we first calculate $Q^\text{vis}= \mathcal{F}^\text{text}\times W^\text{vis}_Q$; $K^\text{vis}= \mathcal{F}^\text{vis}\times W^\text{vis}_K$; $V^\text{vis}= \mathcal{F}^\text{vis}\times W^\text{vis}_V$, then calculate $S^\text{vis}$ as follows:
% \begin{equation}
% \label{eq:image_att}
%     S^\text{vis}=\text{softmax}(\dfrac{Q^\text{vis}\cdot (K^\text{vis})^\mathrm{\top}}{\sqrt{d}})
% \end{equation}
% Similarity, to understand the relationship between the segmentation mask and grasp region features, we calculate attention scores. We first calculate $Q^\text{seg}= \mathcal{F}^\text{vis}\times W^\text{seg}_Q$; $K^\text{seg}= \mathcal{F}^\text{seg}\times W^\text{seg}_K$; $V^\text{seg}= \mathcal{F}^\text{seg}\times W^\text{seg}_V$, then calculate $S^\text{seg}$ as follows:
Similarity, to understand the relationship between segmentation mask and grasp region features, we first obtain $Q^\text{seg}= \mathcal{F}^\text{vis}\times W^\text{seg}_Q$; $K^\text{seg}= \mathcal{F}^\text{seg}\times W^\text{seg}_K$; $V^\text{seg}= \mathcal{F}^\text{seg}\times W^\text{seg}_V$, then calculate $S^\text{seg}$:

% Similarity, for segmentation mask features, we first calculate $Q^\text{seg}= \mathcal{F}^\text{vis}\times W^\text{seg}_Q$; $K^\text{seg}= \mathcal{F}^\text{seg}\times W^\text{seg}_K$; $V^\text{seg}= \mathcal{F}^\text{seg}\times W^\text{seg}_V$, then calculate $S^\text{seg}$ as follows:

\begin{equation}
\label{eq:mask_att}
    S^\text{seg}=\text{softmax}(\dfrac{Q^\text{seg}\cdot (K^\text{seg})^\mathrm{\top}}{\sqrt{d}})
\end{equation}
%\textcolor{blue}{anwser: core calculation for self-attention and cross-attention is very similar. Self-attention: focusing on the sequence embedding itself. In the E.Q1. We just apply to text embedding, so call self-attention. Cross-attention: Instead of just focusing on the meaning of a single sentence, we can also consider how information from another sentence relates to the first sentence's meaning. In the E.q 2,3, for example, with $S^\text{vis}$ , we compute the text embedding related to$\mathcal{F}^\text{vis}$, consider the text embedding as Query. This is similary to the E.Q 3}\textbf{EQ. 2, 3, 4 are similar, why 2 is self-attention and 3 +4 are cross attention?} 

The overall attention outputs can then be computed as $S^\text{text} \times V^\text{text}$, $S^\text{vis} \times V^\text{vis}$ and $S^\text{seg} \times V^\text{seg}$, respectively, which can be applied to the original vectors $\mathcal{F}^\text{text}$, $\mathcal{F}^\text{vis}$ and $\mathcal{F}^\text{seg}$. 
% To suppress the initial cross-modal instability while training due to different embedding spaces, we experimentally observed that disallowing backpropagation of gradients to $\mathcal{F}^\text{text}$, $\mathcal{F}^\text{vis}$ and $\mathcal{F}^\text{seg}$ from the query computation for crossmodality (i.e., $Q^\text{text}$, $Q^\text{vis}$ and $Q^\text{seg}$) led to better model training and performance. %, so we employ that in our model.
The attention can be learned over multiple heads in parallel, as seen in the transformer architecture~\cite{vaswani2017attention}, for added context within the scaling. If the attention layer is splitted into $H$ heads, the output of each head will have a dimension of $d_\text{head} = \frac{d}{H}$. Also, the weight matrices $Q$, $K$ and $V$ will now be of dimensions $d_\text{head} \times d$. The final output of the multi-head attention layer is then given by: 

\begin{equation}
    \text{MultiHeadAttn} = \text{concat}(head_1,...,head_H)W_O
\end{equation}
where $W_O \in \Real^{d\times d}$ are weights to be learned. We use a layer normalization post-scaling to reduce the chances of overfitting. The final output of the scaling results in the following embeddings:
\begin{equation}
    z^\text{text} = \text{LayerNorm}(S^\text{text} \times V^\text{text} + \mathcal{F}^\text{text})
\end{equation}

\begin{equation}
\label{eq:5}
    z^\text{vis} = \text{LayerNorm}(S^\text{vis} \times V^\text{vis} + \mathcal{F}^\text{vis})
\end{equation}

\begin{equation}
    z^\text{seg} = \text{LayerNorm}(S^\text{seg} \times V^\text{seg} + \mathcal{F}^\text{seg})
\end{equation}

An output grasping module that consists of two Multi-Layer Perceptron (MLP) to fused the information of $\{z^\text{text}_1,\cdots,z_M^\text{text}\}$ and $\{z^\text{vis}_1,\cdots,z_M^\text{vis}\}$, then the final MLP layers projects fused features into a set of $M$ grasp scores $\{S^\text{vis}_1,\cdots,S_M^\text{vis}\}$, respectively. The grasp object proposal $G^{i}, i \in M$ with the highest grasping score is selected as the final grasping prediction.

\subsection{Training}
\textbf{Triplet Correspondence Loss.}
The introduced loss function for grasp region features correspondence aims to understand the relationship between grasp regions and those identified in the segmentation mask feature of objects. We formulate this correspondence loss as a triplet loss~\cite{karpathy2015deep,faghri2017vse++,wang2019learning,li2019visual,yang2021sat}: 
\begin{align*}
        \mathcal{L}_{\rm cor} = \sum_{m=1}^M &\left\{ \left[ \alpha- s(z_m^\text{vis},z_m^\text{seg}) + s(z_m^\text{vis},z_{i}^\text{seg}) \right ]_+ \right . \\
+ & \left . \left[ \alpha- s(z_m^\text{vis},z_m^\text{seg}) + s(z_{j}^\text{vis},z_m^\text{seg}) \right ]_+ \right\},
\end{align*}
where $\text{s}(\cdot)$ is the similarity function. We use the inner product over the L2 normalized feature $z^\text{vis}$ and $z^\text{seg}$ as $\text{s}(\cdot)$ in our experiments. $\alpha$ is the margin with a default value of $0.1$. $i,j$ are the index for the hard negatives where $i=\text{argmax}_{i\neq m} s(z_m^\text{vis},z_{i}^\text{seg})$ and $j=\text{argmax}_{j\neq m} s(z_{j}^\text{vis},z_m^\text{seg})$. We determine the grasp correspondence among the attention grasp region features proposals $m$ and the attention segmentation mask features within each input image $I$. 

% \textbf{Segmentation Loss.} 
% In order to obtain meaningful predictions for the object masks, we have utilize a standard cross-entropy segmentation loss function based on binary mask, which measures the discrepancy between the predicted probability distribution and the ground truth labels for each pixel in the image. Formally, the segmentation loss $\mathcal{L}_{\text{seg}}$ can be defined as:
% \begin{equation}
% \mathcal{L}_{\text{seg}} = - \frac{1}{N} \sum_{i=1}^{N} \left( y_i \cdot \log(p_i) + (1 - y_i) \cdot \log(1 - p_i) \right)
% \end{equation}
% where $N$ is the total number of pixels, $y_i$
% represents the ground truth label (1 for foreground, 0 for background) for pixel $i$, and 
% $p_i$ is the predicted probability of pixel 
% $i$ belonging to the foreground class.

% In the total Amodal Segmentation loss $L_seg$:
% % NOTE(mu): Warum ist L_{box} notwendig?
% \begin{equation}
% L_seg = L_{\mathit{AM}} + L_{\mathit{VM}} + L_{\mathit{IVM}},
% \end{equation}
% where \emph{AM, VM, IVM} are abbreviations for amodal, visible, and invisible mask, respectively. In theory, one of the three losses $L_{\mathit{AM}}, L_{\mathit{VM}}$ and $L_{\mathit{IVM}}$ is redundant, as for ground truth we have $\mathit{IVM} = \mathit{AM} - \mathit{VM}$. Nevertheless, adding an additional loss for occlusion masks leads to amodal mask logits and visible mask logits that are on the same scale. Otherwise, consider the case that for a pixel we have high probability for the amodal and the visible mask.

\textbf{Grasp Loss}. The grasp loss function, denoted as $\mathcal{L_{\rm grasp}}$, is informed by previous research on grasp detection~\cite{chu2018real,roiGrasp2019}.  It combines grasp regression and classification losses to serve two main purposes. Firstly, grasp regression loss guarantees accurate grasp localization. Secondly, classification loss assists in accurately identifying successful grasps, crucial for distinguishing effective grasping strategies from ineffective ones.
\begin{equation}
\begin{split}
\mathcal{L_{\rm grasp}} = &-\sum_{i\in \rm Positive}{\log(p_g^{i})}-\sum_{i\in \rm Negative}{\log(p_u^{i})}\\
&+\beta\sum_{i\in \rm Positive}{\text{smoothL1}(G^{i},G^{i}_{gt})}
\end{split}
\end{equation}
where $p_g^{i}$ and $p_u^{i}$ denote probabilities of grasp sample classification into ``graspable" and ``ungraspable", and $G^{i}$ and $G^{i}_{gt}$ denote predicted and ground truth grasps, respectively. In our experiment, $\beta$ is set to $1.4$ to balances the contributions of grasp regression and classification.

Finally, the overall training objective is the combination of both loss terms $\mathcal{L}_{\rm total}$:
\begin{align}
\mathcal{L}_{\rm total}  &=  \mathcal{L}_{\rm grasp}+\lambda_{\text{c}} \mathcal{L}_{\text{cor}} 
%+ \lambda_{\text{s}} \mathcal{L}_{\text{seg}} 
\end{align}
In our experiment, $\lambda_\text{c}$ is set to $0.8$ to balance the loss.
% where $\lambda_\text{c}$ is hyper-parameter to balance loss. In our experiment, $\lambda_\text{c}$ is set to $0.8$.

\section{Experiments} 
\label{Sec:exp}
We first conduct experiments to assess the effectiveness of our proposed method on a large-scale language-driven grasping dataset~\cite{vuong2023grasp}. We further verify our method on real robot grasping experiments. Additionally, we showcase the ablation study of our approach in language-driven grasp detection tasks. Finally, we discuss the encountered challenges and outline open questions for future research.

\begin{figure*}[t]
	\centering
	\includegraphics[width=1.0\linewidth]{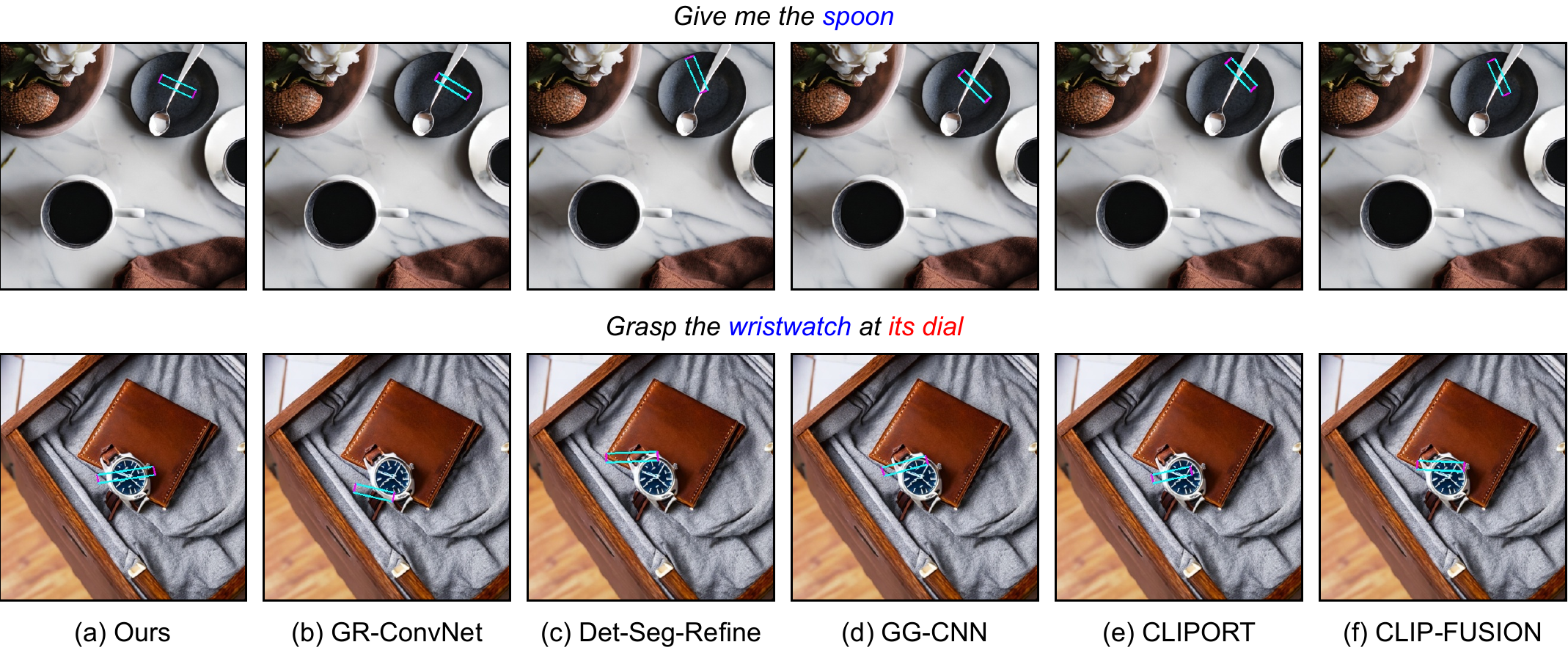}
 \vspace{-2mm}
	\caption{Language-driven grasp detection results.}
	\label{fig:vls_quantitive}
\end{figure*}
\subsection{Experimental Setup}
\textbf{Dataset.} Our experimental setup utilizes the Grasp-Anything dataset~\cite{vuong2023grasp}, a large-scale compilation of grasp data synthesized from foundational models. This dataset boasts diversity and scale, comprising 1M images with textual descriptions and featuring over 3M objects. As in~\cite{vuong2023grasp,zhou2022conditional}, we categorize data into `Seen' and `Unseen' categories, allocating $70\%$ of categories as `Seen' and the remaining $30\%$ as `Unseen'. We also use the harmonic mean ('H') metric to measure overall success rates~\cite{zhou2022conditional}.
%We utilize the Grasp-Anything dataset~\cite{vuong2023grasp} in our experimental setup. This dataset represents a novel large-scale compilation of grasp data synthesized from foundational models. Grasp-Anything stands out due to its remarkable diversity and scale, encompassing 1 million samples accompanied by textual descriptions and featuring over 3 million distinct objects. To facilitate zero-shot learning, we adopt the notion of 'Seen' and 'Unseen' labels~\cite{zhou2022conditional}. Specifically, we allocate $70$\% of these labels based on their frequency as 'Seen' labels, while the remaining $30$\% are designated as `Unseen' labels. In assessing our model's performance, we employ the harmonic mean ('H') as a metric to gauge overall success rates, following the approach outlined in~\cite{zhou2022conditional}. 

\textbf{Evaluation Metrics.} Our principal evaluation metric is the success rate, as defined similarly to~\cite{kumra2020antipodal}. This necessitates that the Intersection over Union (IoU) score of the predicted grasp exceeds $25$\% with the ground truth grasp, and the offset angle is less than $30\degree$. During training, we keep the text encoder and segmentation extractor fixed and then train the rest of the network end-to-end.

\textbf{Baselines.} We compare our method (MaskGrasp) with GR-CNN~\cite{kumra2020antipodal}, Det-Seg-Refine~\cite{ainetter2021end}, GG-CNN~\cite{morrison2018closing}, CLIPORT~\cite{shridhar2022cliport}, and CLIP-Fusion~\cite{xu2023joint}, utilizing either a pretrained CLIP~\cite{radford2021learning} model for text embedding.

% \begin{figure}
% 	\centering
%     \includegraphics[width=1.0\linewidth]{figures/exp/visualize_attention_v39.pdf}
%  \vspace{-2mm}
% 	\caption{The language-driven grasp region attention visualization}
% 	\label{fig:vls}
% \end{figure}

\begin{table}[h]
\caption{\label{table:lgd_results}Language-driven grasp detection results.}
\centering
\renewcommand
\tabcolsep{4.5pt}
% \hspace{1ex}
\vskip 0.1 in
\resizebox{\linewidth}{!}{
\begin{tabular}{@{}lcccccccc@{}}
\toprule
Baseline & Seen & UnSeen & H&\#Params&Inference time\cr 
\midrule
GR-ConvNet~\cite{kumra2020antipodal} + CLIP~\cite{radford2021learning} &0.37 &0.18 &0.24&2.07M&0.022s\\
Det-Seg-Refine~\cite{ainetter2021end} + CLIP~\cite{radford2021learning} &0.30 &0.15 &0.20&1.82M&0.200s \\
GG-CNN~\cite{morrison2018closing} + CLIP~\cite{radford2021learning} &0.12 &0.08 &0.10&1.24M&0.040s \\
CLIPORT~\cite{shridhar2022cliport} &0.36 &0.26 &0.29&10.65M&0.131s\\
CLIP-Fusion~\cite{xu2023joint} &0.40 &0.29 &0.33&13.51M&0.157s\\
% LGD~\cite{}+ BERT~\cite{devlin2018bert} &0.44 &0.38 &0.41 \\
% LGD~\cite{}+ CLIP~\cite{radford2021learning} &0.48 &0.42 &0.45 \\
\midrule
MaskGrasp + BERT~\cite{devlin2018bert} (ours)& 0.47 &0.43 &0.42 &4.91M &0.127s\\
MaskGrasp + CLIP~\cite{radford2021learning} (ours)& \textbf{0.50} &\textbf{0.46} &\textbf{0.45} &4.72M &0.116s\\
\bottomrule
\end{tabular}}
% \vspace{-1ex}
\end{table}
\subsection{Language-driven Grasp Detection Results}
\textbf{Quantitative Results.} Table~\ref{table:lgd_results} shows the comparison between our method and baselines on the Grasp-Anything dataset. Our approach consistently outperforms other baselines with a clear margin in both `Seen' and `Unseen' setups. Moreover, in the `Unseen' setup, our method exhibits significant superiority, surpassing the runner-up, CLIP-Fusion~\cite{xu2023joint} by $0.17$ in success score.

%\textcolor{blue}{This is due to  the segmentation head. But I known that in the previous baseline base An's compare, we maybe not consider the pretrained. So in this case, if we not count inference time of the segment-> the time will be significantly decrease}\textbf{ WHY THE INFERENCE TIME SO SLOW? can we sacrifice accuracy for inference time?}
% Our approach consistently demonstrates superior performance across both datasets by IOU evaluation metrics. Notably, on the Seen datasets, our method outperforms the runner-up model (CLIP-Fusion) by a substantial margin of $10\%$ in IoU. Furthermore, our method exhibits significant superiority over other models on the Unseen dataset, surpassing CLIP-Fusion by $17\%$ in IOU. Moreover, our method achieves substantial performance superiority over others on the H dataset in terms of IOU, with an improvement of $12\%$. 

\textbf{Qualitative Results.} 
Fig.~\ref{fig:vls_quantitive} shows the quantitative evaluation of our method and other baselines. This figure shows that our method produces semantically plausible results, particularly in cluttered scenes where we have occlusions.

%Furthermore, Fig.~\ref{fig:vls} shows the attention map that provides insights into how our model attends to relevant object features during the grasping process. Notably, our method's attention mechanism demonstrates accurate object localization. This visualization reinforces the robustness of our approach in handling complex environmental conditions.

\subsection{Ablation Study}
\begin{figure}[h]
\centering
\includegraphics[width=0.99\linewidth]{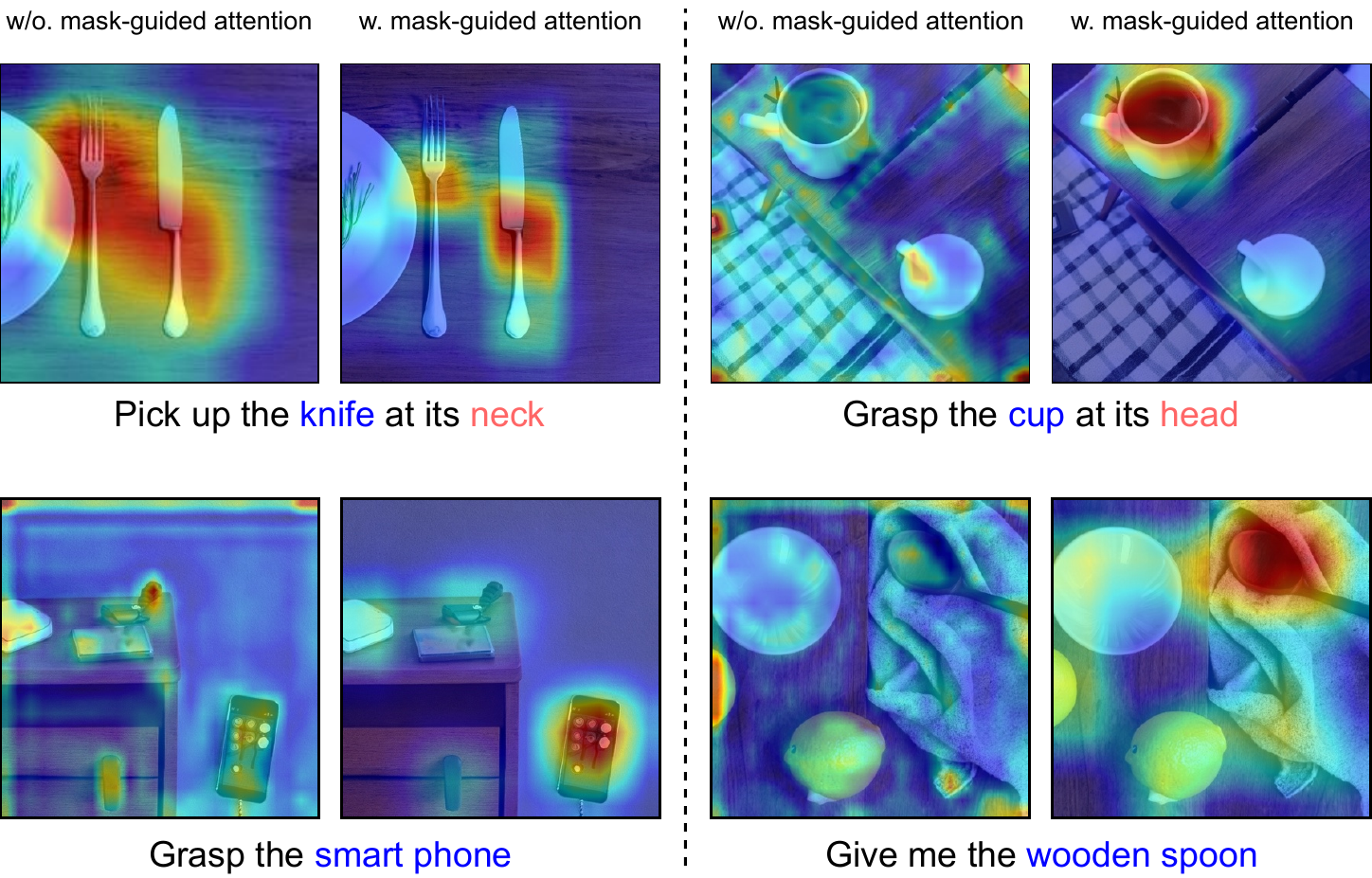}
\vspace{0.05ex}
\caption{The visualization comparison between using and not using our mask-guided attention.}
\label{fig:vls_atten_mask}
\end{figure}

\textbf{Mask-guided Attention Analysis.} To understand how our mask-guided attention performs, we visualize the attention focus under different conditions and compare the model's prioritization of information when provided with both segmentation mask features and text instructions versus text alone. Our results, depicted in Fig.~\ref{fig:vls_atten_mask}, indicate that our method concentrates attention on the target object more effectively when segmentation mask features are available, suggesting that these features guide the model's focus towards crucial regions and facilitate the extraction of richer contextual information, thereby improving grasp performance.

\begin{figure} [t]
\centering
\includegraphics[width=0.99\linewidth]{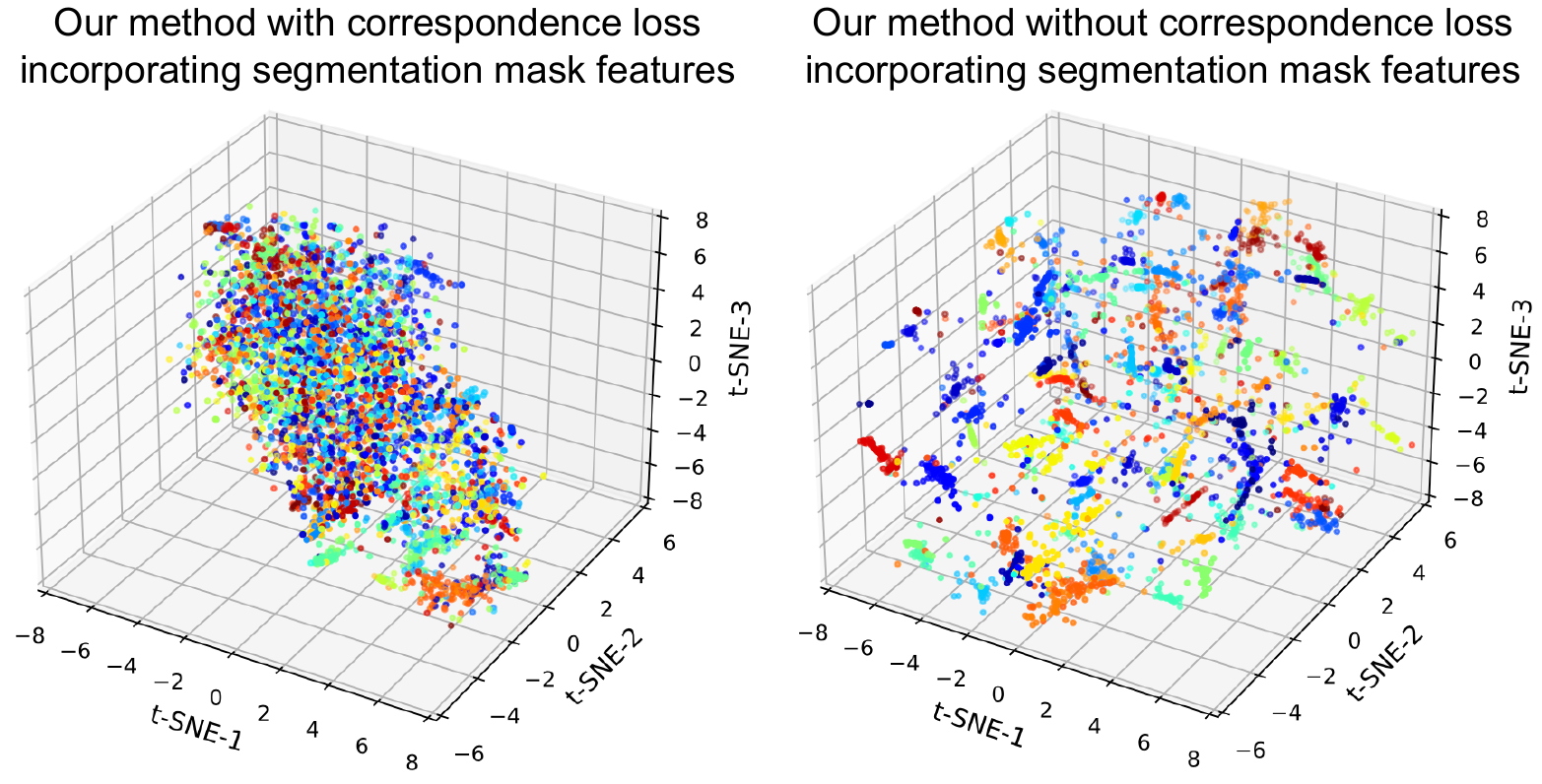}
\vspace{1ex}
\caption{\textbf{t-SNE visualization of the grasp object feature representations}. We apply t-SNE to cluster the grasp object feature representations $z^{vis}$ of Equation~\ref{eq:5} when using and not using the correspondence loss with mask feature objects in our method.}
\label{fig:vls_tsn}
\end{figure}

\textbf{Effectiveness of Segmentation Features.} To assess the importance of segmentation mask features for grasping, we compared our model's performance with and without them. Using t-SNE~\cite{van2008visualizing}, we cluster the grasp region feature representations $z^{vis}$ (Equation~\ref{eq:5}) under both conditions. Fig.~\ref{fig:vls_tsn} illustrates that without correspondence loss with mask-guided features, the decision boundaries for most grasp region features are indistinct and challenging to discern during training. Conversely, applying correspondence loss with mask-guided features enhances both the accuracy and learned grasp region features of the network. These findings, supported by detailed results in Table~\ref{table:loss}, underscore the positive impact of integrating mask-guided features and the corresponding loss function on grasping performance.

\begin{table}[t]
\caption{\label{table:loss} Ablation Study.}
\vspace{-2ex}
\centering
\renewcommand
\tabcolsep{4.5pt}
\hspace{1ex}
\vskip 0.1 in
\resizebox{\linewidth}{!}{
\begin{tabular}{@{}lcccccc@{}}
\toprule
Baseline & Seen & UnSeen & H\cr 
\midrule
Ours w/o segmentation mask & 0.432 &0.314 &0.349\\
Ours w/o correspondence loss &0.483 &0.429 &0.447\\
Ours & \textbf{0.500} &\textbf{0.460} &\textbf{0.451} \\
\bottomrule
\end{tabular}}
%\vspace{-3ex}
\end{table}

%To assess the importance of segmentation object features for grasping, we compared our model's performance with and without them. Specifically, we apply t-SNE~\cite{van2008visualizing} to cluster the grasp object feature representations $z^{vis}$ of Equation~\ref{eq:5} when using and not using the correspondence loss with mask feature objects in our method. Fig.~\ref{fig:vls_tsn} shows that without correspondence loss with mask-guided features, the decision boundaries for most of the grasp object features are obscure and difficult to distinguish during the training. On the other hand, applying correspondence loss with mask-guided features increases both the accuracy and learned grasp object features of the network. These findings, along with detailed results in Table~\ref{table:lgd_results}, highlight the positive impact of incorporating mask-guided features and the corresponding loss function on grasping performance.

\begin{figure}[]
\centering
\includegraphics[width=0.99\linewidth]{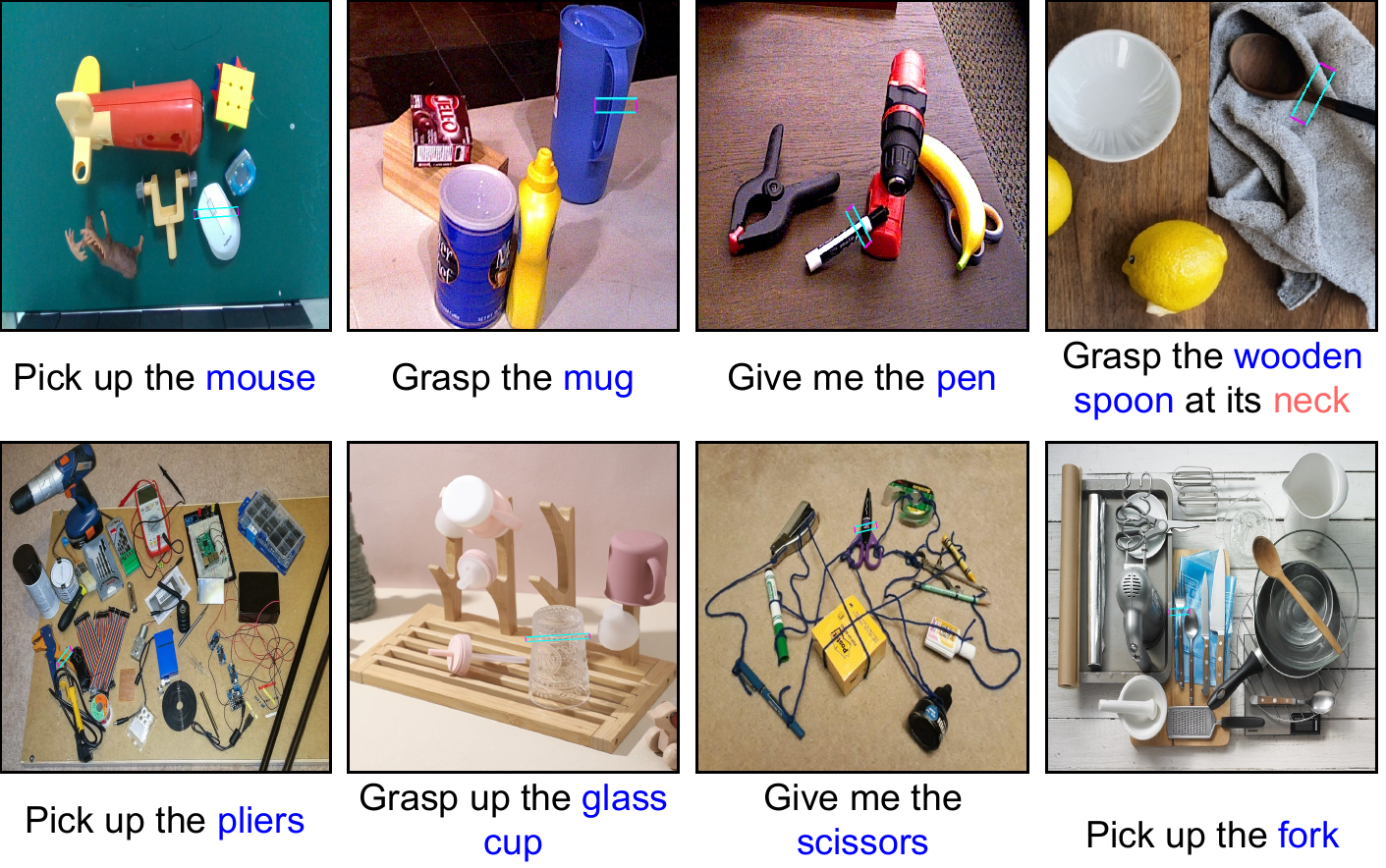}
\vspace{-1mm}
\caption{In the wild detection results. Top row images are from
GraspNet~\cite{xiang2017posecnn}, YCB-Video~\cite{fang2020graspnet} datasets; bottom
row shows internet images.}
\label{fig:inthewild}
\end{figure}

\textbf{In the Wild Detection.} Fig.~\ref{fig:inthewild} showcases visualizations produced by our method, trained solely on the Grasp-Anything dataset, when applied to random internet images and other dataset images. These results demonstrate our model's robust generalization to real-world images, despite being trained solely on synthetic data from Grasp-Anything, without real image inputs.

\begin{table}[t]
    \centering
    \caption{\label{table: real-robot-exp} Robotic language-driven grasp detection results}
    \vspace{2ex}
    \renewcommand
\tabcolsep{4pt}
\hspace{1ex}
    \begin{tabular}{@{}rcc@{}}
\toprule
Baseline & Single &  Cluttered\cr 
\midrule
GR-ConvNet~\cite{kumra2020antipodal} + CLIP~\cite{radford2021learning}  &0.33  & 0.30\\ 
Det-Seg-Refine~\cite{ainetter2021end} + CLIP~\cite{radford2021learning} &0.30  & 0.23\\ 
GG-CNN~\cite{morrison2018closing} + CLIP~\cite{radford2021learning} &0.10  & 0.07 \\
CLIPORT~\cite{shridhar2022cliport} &0.27 & 0.30 \\
CLIP-Fusion~\cite{xu2023joint} & 0.40 & 0.40 \\
\midrule
MaskGrasp (ours) &  \textbf{0.43} & \textbf{0.42} \\
\bottomrule
\end{tabular}
\end{table}

\subsection{Robotic Experiment}
\begin{figure}
\centering
\includegraphics[width=1.\linewidth]{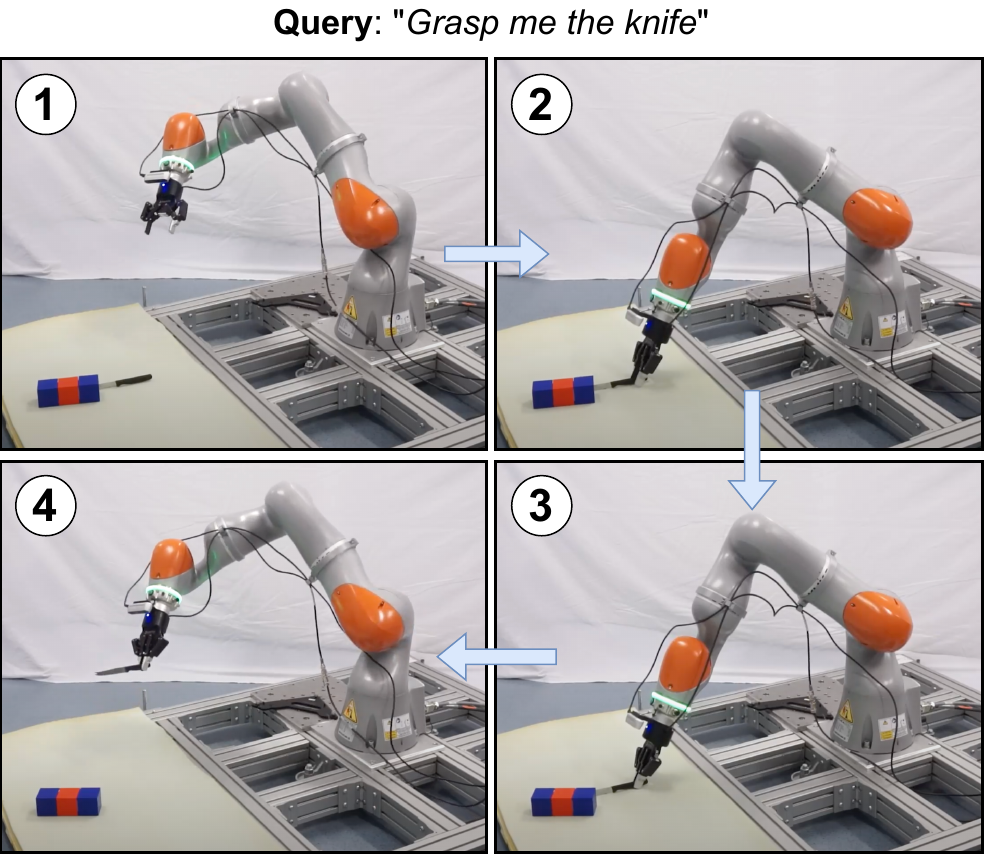}
\vspace{-2mm}
\caption{The robotic experiment setup and sequence of grasping actions.}
\label{fig: robot demonstration}
\end{figure}

In Fig.~\ref{fig: robot demonstration}, we showcase our robotic evaluation using a KUKA robot. Grasp detection, alongside other methods listed in Table~\ref{table: real-robot-exp} of the main paper, is evaluated using depth images from an Intel RealSense D435i depth camera following methodology in~\cite{kumra2020antipodal}. Our proposed method infers 4-DoF grasp poses, transformed into 6 DoF poses under the assumption of flat surface objects. Trajectory optimization detailed in~\cite{beck2023singularity,vu2023machine} guides the robot to target poses. The setup involves two computers: PC1 handles real-time control, camera, and gripper, while PC2 runs ROS on Ubuntu Noetic 20.04, communicating with the robot via EtherCAT protocol. PC2, equipped with an NVIDIA RTX 3080 Ti graphics card, manages the inference process. We assess performance across single-object and cluttered scenarios with a diverse set of real-world objects, repeating each experiment for all methods $30$ times to ensure reliability.

Our proposed method, incorporating mask-guided attention guidance,  outperforms other baselines, as shown in Table~\ref{table: real-robot-exp}. Remarkably, despite being trained solely on Grasp-Anything, a synthetic dataset generated by foundational models, it performs well on real-world objects.
% Our proposed method, incorporating Mask-guided Attention guidance, outperforms other baselines (Table~\ref{table: real-robot-exp}). Notably, even though our model is primarily trained on the synthetic Grasp-Anything dataset, which is exclusively generated by foundational models, it still showcases remarkable efficacy when applied to real-world objects
% , underscoring its robustness and versatility in practical scenarios.
% Our method, incorporating Mask-guided Attention, outperforms other baselines (Table~\ref{table: real-robot-exp}). Notably, trained only on Grasp-Anything, a synthetic dataset, it performs well on real-world objects.

\section{Discussion}\label{Sec:con}
\begin{figure} [t]
\centering
\includegraphics[width=0.99\linewidth]{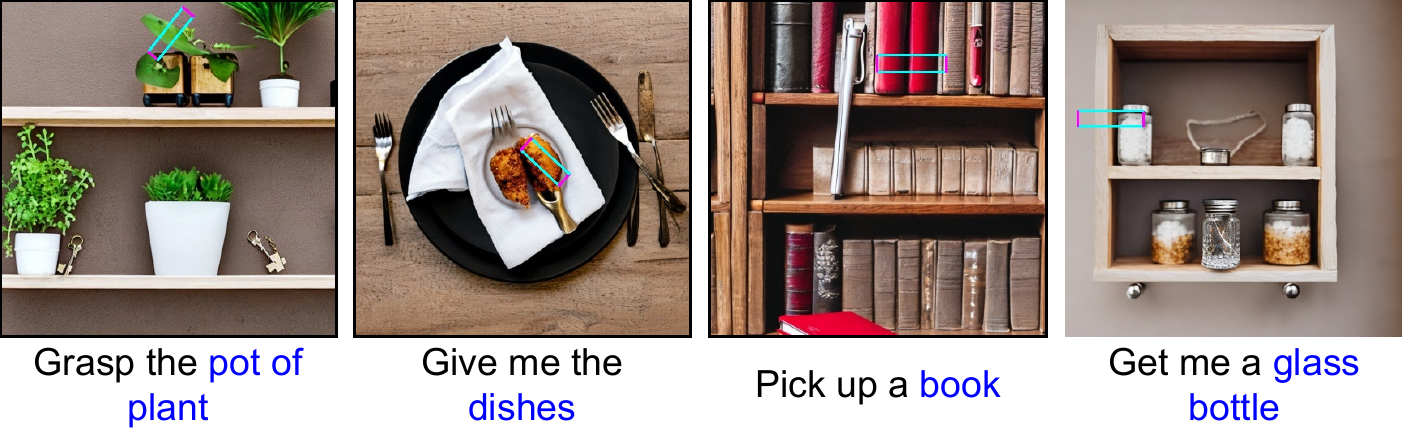}
\vspace{-2mm}
\caption{Failure cases of our method.}
\label{fig:falures}
\end{figure}

\textbf{Limitation.} %Despite notable improvements in generalization, our method faces challenges in scenes with complex semantic object relationships and intricate geometries. Ambiguities arising from contextual cues can hinder accurate object association with language-driven instructions, as shown in Fig.~\ref{fig:falures}. Further refinements may be needed to address these scenarios effectively.
Despite significantly improving generalization capabilities, our method faces challenges when handling scenes with complex semantic object relationships and intricate geometries. Ambiguities arising from contextual cues can impede the accurate association of objects with language-driven instructions, as demonstrated by some failure cases in Fig.~\ref{fig:falures}. Further refinement may be required to effectively address such scenarios.
% \vspace{-2mm}

\textbf{Future work.}  While our study represents a new method in language-driven grasp detection with mask-guided attention, several promising avenues for future research warrant exploration. Further investigation into nuanced attention mechanisms and refinement of handling complex semantic relationships between objects and language instructions are essential~\cite{chen2024training}. Additionally, the integration of reinforcement learning techniques offers the potential for developing adaptive grasp strategies tailored to specific tasks and environments~\cite{nasiriany2022augmenting}. These directions hold promise for enhancing the capabilities and applicability of language-driven robotic grasping systems in real-world scenarios. 

\section{Conclusion}
We introduce a mask-guided attention mechanism to improve multimodal integration for the language-driven grasp detection task. By combining a transformer with segmentation mask-conditioned attention, our method effectively integrates visual and textual information, enhancing grasping accuracy and adaptability. This mechanism prioritizes crucial regions in both modalities, resulting in improved performance. The intensive experiments show that our method outperforms other baselines by a clear margin in vision-based benchmarks and real-world robotic grasping experiments. Our source code and trained model will be made publicly available to facilitate future studies. 

% We have presented a mask-guided attention mechanism to
% enhance multimodal integration for the language-driven
% grasp detection task. Our method combines a transformer with segmentation mask-conditioned attention, allowing it to effectively fuse visual and textual information for accurate and context-sensitive grasping. This attention mechanism focuses on crucial regions in both modalities, ultimately improving grasping performance and adaptability. These enhancements hold substantial promise to apply our proposed method to different robotic applications. Our source code and trained model will be made publicly available. 
% \lipsum[1]

%\section*{Acknowledgment}
%\addcontentsline{toc}{section}{Acknowledgment}
%\lipsum[1]

\bibliographystyle{class/IEEEtran}
\bibliography{class/IEEEabrv,class/reference}
   
\end{document}